\documentclass{article}

 \PassOptionsToPackage{numbers}{natbib}
\usepackage[final]{nips_2017}

\usepackage[utf8]{inputenc} 
\usepackage[T1]{fontenc}    
\usepackage{hyperref}       
\usepackage{url}            
\usepackage{booktabs}       
\usepackage{amsfonts}       
\usepackage{nicefrac}       
\usepackage{microtype}      
\usepackage[pdftex]{graphicx}
\usepackage{listings}
\usepackage{hhline}
\usepackage[small,compact]{titlesec} 
\usepackage[linesnumbered,ruled]{algorithm2e}
\SetAlFnt{\small}
\SetAlCapFnt{\small}
\SetAlCapNameFnt{\small}

\usepackage{comment}

\makeatletter 
\AtEndDocument{%
\def\cb@checkPdfxy#1#2#3#4#5{%
\cb@@findpdfpoint{#1}{#2}%
\ifdim#3sp=\cb@pdfx\relax      
\ifdim#4sp=\cb@pdfy\relax      
\ifdim#5=\cb@pdfz\relax
\else
\cb@error
\fi
\else
\cb@error
\fi
\else
\cb@error
\fi
}}
\makeatother 

\usepackage{color} 
\usepackage[color]{changebar}
\nochangebars
\cbcolor{red}
\newcommand{\Omit}[1]{}

\newcommand{\QED}{\mbox{$\Box$}}
\newcounter{my_counter}

\newcommand{\MyRoman}[1]%
        {\setcounter{my_counter}{#1}\roman{my_counter}}
\newcommand{\PMyRoman}[1]%
        {\rm (\MyRoman{#1})}
\newtheorem{MyTheorem}{Theorem}[section]

{\par\noindent{\bf Theorem~\ref{The:#1}\/}\begin{em}}%
{\end{em}$\QED$}
\newtheorem{Proposition}[MyTheorem]{Proposition}

{\par\noindent{\bf Proposition~\ref{Pro:#1}\/}\begin{em}}%
{\end{em}$\QED$}
\newtheorem{Lemma}[MyTheorem]{Lemma}

{\par\noindent{\bf Lemma~\ref{Lem:#1}\/}\begin{em}}%
{\end{em}$\QED$}
{\par\noindent\begin{rm}{\em Claim\/}:~#1}%
{\end{rm}}

\newtheorem{Corollary}[MyTheorem]{Corollary}

\newtheorem{Conjecture}[MyTheorem]{Conjecture}

\newtheorem{Observation}[MyTheorem]{Observation}

\newtheorem{Definition}[MyTheorem]{Definition}
\newenvironment{Def}{\begin{Definition}}{$\QED$\end{Definition}}
\newtheorem{SDefinition}[MyTheorem]{Strawman Definition}

\newtheorem{Desideratum}[MyTheorem]{Desideratum}

\newtheorem{Example}[MyTheorem]{Example}
\newenvironment{Exa}{\begin{Example}}{$\QED$\end{Example}}
                        {$\QED$\end{quote}}

\usepackage{enumitem}
\setlist{nosep,leftmargin=*}

\title{Neural Attribute Machines for Program Generation\thanks{Supported, in part,
  by a gift from Rajiv and Ritu Batra;
  by NSF under grant CCF-1162076;
  by AFRL under DARPA MUSE award FA8750-14-2-0270 and DARPA STAC award FA8750-15-C-0082;
  and by the UW-Madison Office of the Vice Chancellor for
  Research and Graduate Education with funding from the Wisconsin Alumni
  Research Foundation.
  Any opinions, findings, and conclusions or recommendations
  expressed in this publication are those of the authors,
  and do not necessarily reflect the views of the sponsoring
  agencies.}
}

\author{
  Matthew Amodio \\
  Computer Sciences Department \\
  University of Wisconsin \\
  \texttt{mamodio@cs.wisc.edu} \\
  \And
  Swarat Chaudhuri \\
  Department of Computer Science \\
  Rice University \\
  \texttt{swarat@rice.edu} \\
  \And
  Thomas Reps \\
  Comp.\ Sci.\ Dept., Univ.\ of Wisconsin \\
  \& GrammaTech, Inc. \\
  \texttt{reps@cs.wisc.edu} \\
}

\begin{document}

\maketitle

\begin{abstract}
Recurrent neural networks have achieved remarkable success at
generating sequences with complex structures, thanks to advances that
include richer embeddings of input and cures for vanishing
gradients. Trained only on sequences from a known grammar, though,
they can still struggle to learn rules and constraints of the
grammar.
Neural Attribute Machines (NAMs) are equipped with a
logical machine that represents the underlying grammar, which is used
to teach the constraints to the neural machine by (i) augmenting the
input sequence, and (ii) optimizing a custom loss function. Unlike
traditional RNNs, NAMs are exposed to the grammar, as well as
samples from the language of the grammar. During generation, NAMs
make significantly fewer violations of the constraints of the
underlying grammar than RNNs trained only on samples from the language
of the grammar.
\end{abstract}

\section{Introduction}

Neural networks have been applied successfully to many generative
modeling tasks, from images with pixel-level detail
\cite{oord2016pixel} to strokes corresponding to crude sketches
\cite{ha2017neural} to natural language in automated responses to user
questions \cite{vinyals2015show}. Less extensively studied have been
neural models for generation of highly structured artifacts, for
example {\em the source code of programs}. Program generation has many
potential applications, including automatically testing programming
tools~\cite{holler2012fuzzing} and assisting humans as they solve
programming tasks~\cite{hindle2012naturalness,raychev2014code}.
\begin{changebar}
However, a fundamental
difficulty in this problem domain is that to be acceptable to a compiler, a
program must satisfy a rich set of constraints such as ``never use
undeclared variables'' or ``only use variables in a type-safe way''.
\end{changebar}
Learning such constraints automatically from data is a difficult task.

In this paper, we present a new generative model, called \emph{Neural
Attribute Machines} (NAMs), for 
programs that satisfy constraints like the above. The key insight of
our approach is that the constraints enforced by a programming
language are known in full detail \textit{a priori}. Accordingly, they
can be incorporated into training, and we propose a framework for
doing so.
\begin{changebar}
We demonstrate that this approach has significant benefits: training
existing architectures on samples that unfailingly abide by a
constraint still produces a generative model that often violates the
constraint;
in contrast, the NAM model
significantly outperforms these models at sampling from the space of
constrained programs. 
\end{changebar}


We use the formalism of \emph{attribute grammars} \cite{MST:Knuth68}
as the language for expressing rich structural constraints over
programs.
\begin{changebar}
Our model composes such a grammar with a recurrent neural network
(RNN) that generates the nodes of a program's abstract-syntax tree,
and uses it to constrain the output of the RNN at each point in time.
\end{changebar}
Our training framework builds off of the observation that in the setting
of generating constrained samples, there is a correct prediction, and
then there are \textit{two categories} of incorrect
predictions. Incorrect predictions that are nevertheless
\textit{legal} under the constraint are more desirable than
predictions that \textit{violate} the constraint. The NAM addresses
this multifaceted problem in two ways. First, it augments the input
sequence with a fixed-length representation of a \emph{structural
context} that the attribute grammar uses to check constraints. The
context provides information that the current input sequence is just
one particular instantiation of a more general structural
constraint. Second, NAMs optimize a three-valued loss function that
penalizes correct, incorrect-but-legal, and incorrect-and-illegal
predictions differently.

The main contributions of this paper can be summarized as follows:
\begin{itemize}
  \item
    We present a new neural network and logical architecture that
    incorporates background knowledge in the form of attribute-grammar
    constraints.
  \item
    We give a framework to train this new architecture that uses a
    three-valued loss function to penalize correct, incorrect-but-legal,
    and incorrect-and-illegal predictions differently.
  \item
    Our experiments show the difficulties existing neural models have with
    learning in constrained domains, and the advantages of using the
    proposed framework.
\end{itemize}


\section{Methodology}

\subsection{Background on Attribute Grammars}

\begin{Def}\label{De:AttributeGrammar}
An \textbf{attribute grammar} (AG) is a context-free grammar extended by
attaching \textbf{attributes} to the terminal and nonterminal symbols
of the grammar, and by supplying \textbf{attribute equations} to
define attribute values \cite{MST:Knuth68}.
\begin{changebar}[6pt]
Each production can also be equipped with an \textbf{attribute
constraint} to specify that some relationship must hold among the
values of the production's attributes.
\end{changebar}

In every production $X_0 \rightarrow X_1, \ldots, X_k$, each $X_i$
denotes an \textbf{occurrence} of one of the grammar symbols;
associated with each such symbol occurrence is a set of
\textbf{attribute occurrences} corresponding to the symbol's
attributes.

The attributes of a symbol $X$, denoted by $A(X)$, are divided into
two disjoint classes:
\textbf{synthesized} attributes and \textbf{inherited} attributes.
A production's \textbf{output} attributes are the synthesized-attribute
occurrences of the left-hand-side nonterminal plus the
inherited-attribute occurrences of the right-hand-side symbols;
its \textbf{input} attributes are the inherited-attribute occurrences
of the left-hand-side nonterminal plus the synthesized-attribute
occurrences of the right-hand-side symbols.

Each production has a set of attribute equations, each of which defines
the value of one of the production's output attributes as a function
of the production's input attributes.
We assume that the terminal symbols of the grammar have no synthesized
attributes, and that the root symbol of the grammar has no inherited
attributes.
Noncircularity is a decidable property of AGs
\cite{MST:Knuth71}, and hence we can assume that no derivation tree
exists in which an attribute instance is defined transitively in terms
of itself.

An AG is \textbf{$L$-attributed} \cite{JCSS:LRS74} if, in
each production $X_0 \rightarrow X_1, \ldots, X_k$, the attribute
equation for each inherited-attribute occurrence of a right-hand-side
symbol $X_i$ only depends on
(i) inherited-attribute occurrences of $X_0$, and
(ii) synthesized-attribute occurrences of $X_1, \ldots, X_{i-1}$.
\end{Def}

\begin{changebar}[6pt]
With reasonable assumptions about the computational power of attribute
equations and attribute constraints, $L$-attributed AGs capture the
class NP \cite{IC:EPS88} (modulo a technical ``padding'' adjustment).
\end{changebar}

\begin{figure}
\centering
\begin{tabular}{c@{\hspace{5.0ex}}c}
  {\small
  {\rm
  \begin{tabular}{rcl}
    numeral & : & Numeral \{ bits.positionIn = 0; \}; \\
    bits    & : & Pair \{ \\
            &   & ~~~~bits\$2.positionIn = bits\$1.positionIn; \\
            &   & ~~~~bits\$3.positionIn = bits\$2.positionOut; \\
            &   & ~~~~bits\$1.positionOut = bits\$3.positionOut; \\
            &   & \} \\
            & | & Zero, One \{ \\
            &   & ~~~~bits.positionOut = bits.positionIn + 1; \} \\
            &   & \} \\
            & ; & \\
  \end{tabular}
  }}
  &
   \begin{minipage}{.32\textwidth}
     \includegraphics[width=1\linewidth]{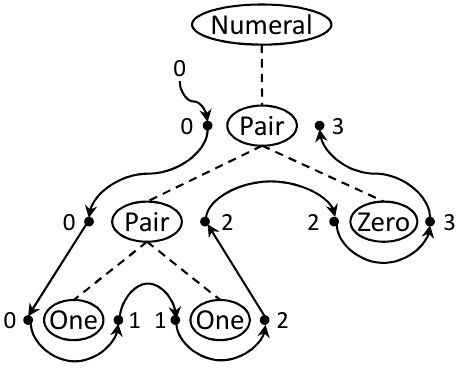}
   \end{minipage}
  \\
  (a) & (b)
\end{tabular}
\caption{(a) Attribution rules for left-to-right threading.
         (b) A derivation tree, which illustrates the left-to-right
             pattern of dependences among attribute instances:
             at each node, the attributes positionIn and positionOut
             are shown on the left and right, respectively.
        }
\label{fig:LRThreadingRules}
\end{figure}

\begin{Exa}\label{Exa:LRThreading}
To illustrate $L$-attributed AGs, we use a simple language of binary
numerals.
The abstract-syntax trees for binary numerals are defined using
the following operator/operand declarations:\footnote{
  The notation used above is a variant of context-free grammars in
  which the operator names (Numeral, Pair, Zero, and One) serve to
  identify the productions uniquely.
  For example, the declaration ``\textrm{numeral: Numeral(bits);}''
  is the analog of the production
  ``$\textrm{numeral} \rightarrow \textrm{bits}$.''
  (The notation is adapted from the Synthesizer Generator \cite{Book:RT88}.)
}

\noindent
{\rm
\begin{tabular}{r@{\hspace{1.0ex}}c@{\hspace{1.0ex}}l@{\hspace{8.0ex}}r@{\hspace{1.0ex}}c@{\hspace{1.0ex}}l}
  \qquad\qquad
  numeral & : & Numeral(bits); & bits & : & Pair(bits bits)~~~|~~~Zero()~~~|~~~One();
\end{tabular}
}

\noindent
Two integer-valued attributes---''positionIn'' and
``positionOut''---will be used to determine a bit's position in a
numeral:

\noindent
{\rm
\begin{tabular}{r@{\hspace{1.0ex}}c@{\hspace{1.0ex}}l}
  \qquad\qquad\qquad
  bits & \{  & inherited int positionIn; ~~~~synthesized int positionOut; \};
\end{tabular}
}

\noindent
These attributes are used to define a left-to-right pattern of
information flow through each derivation tree (also known as
``left-to-right threading'').
\begin{changebar}
In particular, with the attribute equations given in Fig.\ \ref{fig:LRThreadingRules}(a),
at each leaf of the tree, the value of bits.positionOut is the
position of the bit that the leaf represents with respect to the left
end of the numeral
\end{changebar}
(where the leftmost bit is considered to be at position 1).\footnote{
  In the attribute equations for a production such as
  ``bits: Pair(bits bits);''
  we distinguish between the three different occurrences of
  nonterminal ``bits'' via the symbols ``bits\$1,'' ``bits\$2,'' and
  ``bits\$3,'' which denote the leftmost occurrence, the
  next-to-leftmost occurrence, etc.
  (In this case, the leftmost occurrence is the left-hand-side
  occurrence.)
}
Fig.\ \ref{fig:LRThreadingRules}(b) shows the left-to-right pattern of
dependences among attribute dependences in a six-node tree.
\end{Exa}

Our ultimate goal is the creation of new trees \textit{de novo}, with
generation proceeding top-down, left-to-right.
The latter characteristic motivated the choice that constraints be
expressible using an $L$-attributed AG, because they
propagate information left-to-right in an AST.

$L$-attributed AGs also offer substantially increased
expressive power over context-free grammars---in particular, an
$L$-attributed AG $G$ can express \emph{non-context-free
constraints} on the trees in the language ${\cal L}(G)$.
There are a large number of such constraints involved in any grammar
that produces only programs that pass a C compiler.
\begin{changebar}
In the study presented in Section \ref{ssec:Experiments}, we experimented
with two constraints in isolation:
\end{changebar}
\begin{description}
  \item [Declared-variable constraint:]
        Each use of a variable must be preceded by a declaration of the variable.
  \item [Typesafe-variable constraint:]
        Each use of a variable must satisfy the type requirements of
        the position of the variable in the tree.
\end{description}
\begin{changebar}
By working with a corpus of C programs that all compile, all of the
training examples satisfy both constraints.
\end{changebar}

\subsection{From an AST to a sequence}
\begin{figure}

\begin{minipage}{.49\textwidth}
	\centering
	\includegraphics[width=.8\linewidth]{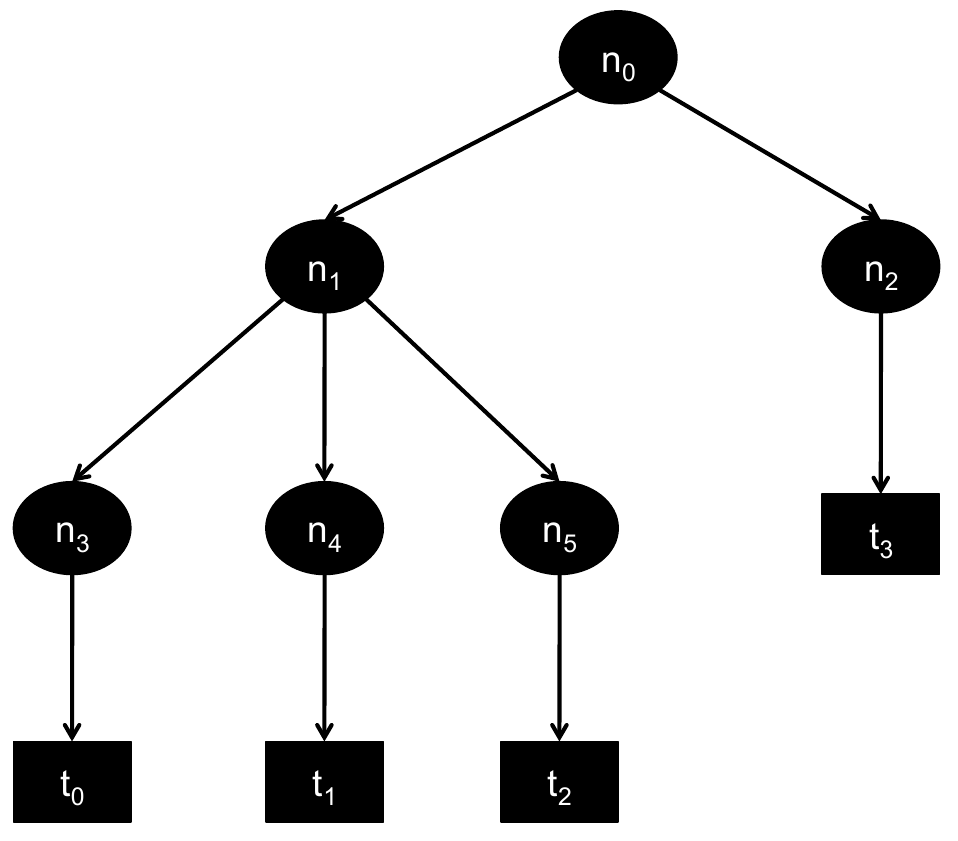}
\end{minipage}%
\hfill%
\begin{minipage}{.49\textwidth}
	\centering
	\includegraphics[width=.8\linewidth]{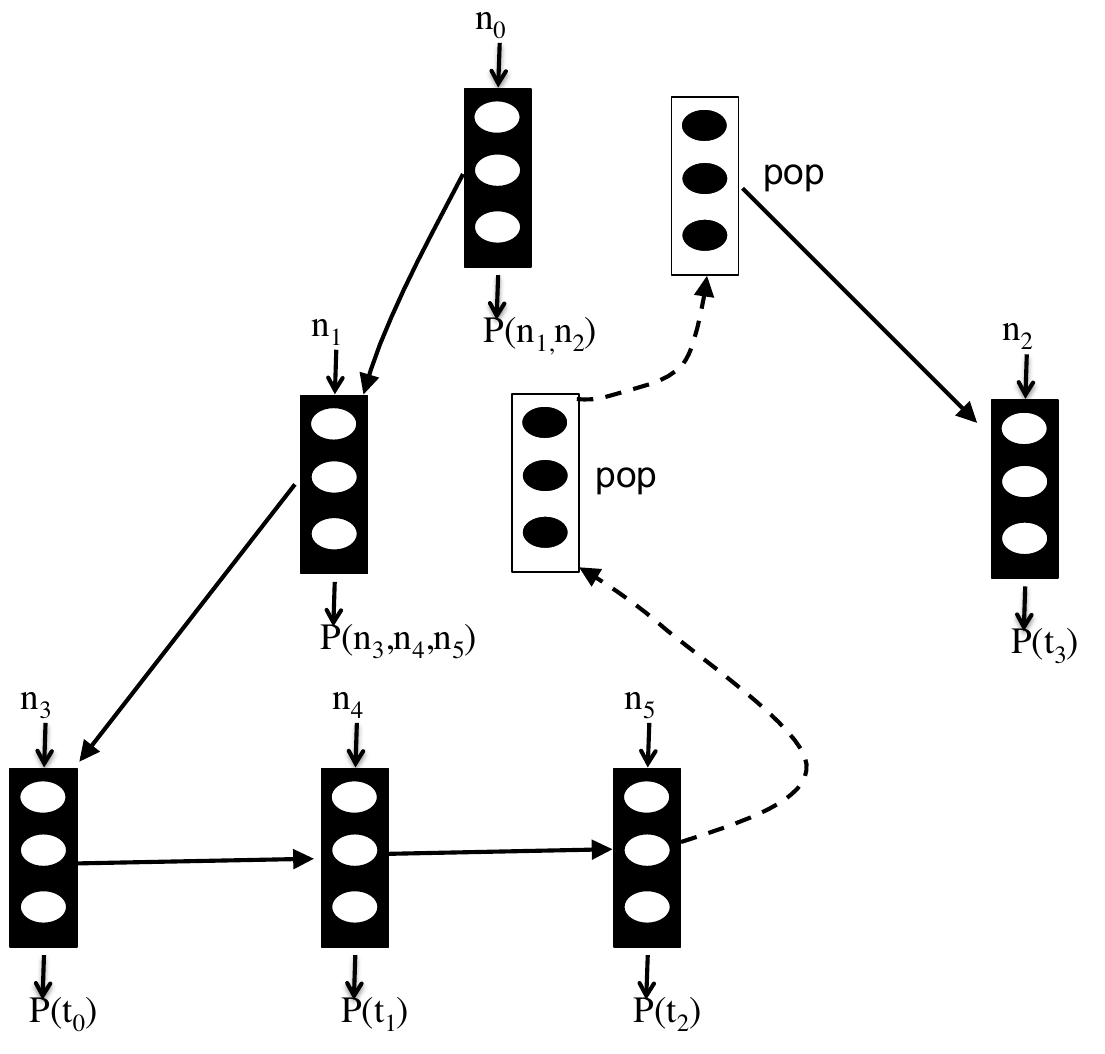}
\end{minipage}
\caption{(a) An AST $T$ and (b) a run of a NAM that generates $T$}
\label{fig:AST}

\end{figure}

While other attempts at learning models that can be used to generate
trees include performing convolutions over nodes in binary trees
\cite{mou2015discriminative} or stacking multiple RNNs in fixed
directions \cite{shuai2016dag}, a natural paradigm to adopt for
presenting a tree to a neural model is a single-traversal, top-down,
left-to-right sequentialization of the tree.
\begin{changebar}
An AST $T$ is represented by a depth-first sequence of pairs:
each pair $(n_i, p_i)$, for $1 \leq i \leq |\textit{Nodes(T)}|$,
consists of a nonterminal $n_i \in \textit{N}$ and a production
$p_i \in P$ that has $n_i$ on the left-hand side.
\end{changebar}
Depth information in the tree is conveyed by interspersing
\textit{pop} indicators when moving up the tree.
As a preprocessing step, all variable names are
aliased by their order of use in the program so that the $i^{th}$
distinct variable used is named Var$i$.
This approach prevents difficulties associated with rarely used
token names, and does not lose any meaningful information about
structural constraints.
Fig. \ref{fig:AST} depicts a trivial example to illustrate the process.

\begin{changebar}
Let $|N|$ and $|P|$ denote the input and output dimensions, respectively.
If $n$ is a nonterminal in $N$, we use $P_n$ to denote the subset of the
productions $P$ with $n$ on the left-hand side.
$P_n$ is the set of legal outputs under the context-free constraint.
The two context-sensitive constraints considered here further narrow
the set of legal outputs.
We use $P_{c_d}$ and $P_{c_t}$ to denote the set of legal outputs at some
(unspecified) point in the linearized tree, under the declared-variable
constraint $c_d$ and the type constraint $c_t$, respectively.
Let $c$ be either $c_d$ or $c_t$;
at a prediction step at an instance of nonterminal $n$, the possible output can be
partitioned into three sets: $\{p_i\}$, $P_c - \{p_i\}$, and $P_n - P_c$.
\end{changebar}

\subsection{Challenges}
Large and variably sized AST sequences present several problems for traditional neural sequence-learners. The first is the existence of distant non-local dependences, such as a declaration of variable $v$ near the beginning of a file that might be many hundreds of steps away before the RNN needs to predict the use of $v$. Another is the existence of complex relationships between nodes that are difficult to express in a linear sequentialization, like when the distinction between a great-grandparent and a great-uncle node is important. Third, due to only approximating the sequence distribution, it is very likely that while generating large ASTs under randomized sampling, novel contexts will be encountered. In such a case, there would not be any information explicitly contained in the training set that can guide further generation after entering ``uncharted territory.'' With other approaches, one relies on the ability of the learner to generalize from the training set. However, imperfect learning of the constraints of \textit{L} could cause poor generalization. Because NAMs are able to learn from the constraints, they have the potential to generalize based on the constraints, and hence have the potential to perform much better when they enter uncharted territory. The new idea pursued here is to leverage constraints that are defined unambiguously even in novel situations. If (an approximation to) the constraints can be learned, they will provide additional guidance about what to do in these situations.

\subsection{NAM}
NAMs are equipped with a deterministic automaton, referred to hereafter as the \textit{logical machine}. The logical machine provides assistance to the NAM in two ways: it augments the input vector with a fixed-length vector that represents the context, and it imposes its knowledge of the output partitions $\{p_i\}$, $P_c - \{p_i\}$, and $P_n - P_c$ to add an extra loss term for constraint-violating predictions.

\paragraph{Augmented input.}
The logical machine outputs a fixed-length binary vector $C$ that represents the context of the current node with respect to the constraint desired. All variable names are known from the grammar $G$, and each corresponds to one production rule $p \in P$. Let $p_{v_i}$ be the production rule that chooses variable name $v_i$ and $P_v$ be the collection of all production rules for variables. For the context vector $C_{d}$ for the declared-variable constraint, which is binary and of length $|P_v|$, the $1$-valued entries are the $p_{v_i}$ positions of declared variables\Omit{, which the logical machine can supply unambiguously}.

For the context vector $C_{t}$ for the typesafe-variables constraint\Omit{, converting the constraint into a fixed-length representation is a little less straightforward. E}, each variable and type combination $(v_i, t_i)$ has an entry in $C_{t}$, which is $1$ if $v_i$ is of type $t_i$. There is then one additional entry for each type $t_i$, which is $1$ if the current prediction must be of type $t_i$.

Developing fixed-length context vectors that are informative for the constraint at hand could be daunting for constraints that are more conceptually complex, but even these simple representations had a profound positive impact on the model learned. Different representations of the same constraint were not tried\Omit{, these being simply the authors' first attempts.}. An interesting direction for future work would be to test how robust the NAM framework is to different ways of building the context vector.

\paragraph{Three-level loss function.}
In addition to being presented with augmented input by the logical machine, NAMs are also trained with additional reinforcement from the logical machine. The standard cross-entropy loss function that measures the distance between the model's predicted probability distribution $\hat{y}$ and the true observation $y$ suffers from an undesirable consequence in the setting of learning to generate constrained sequences. The one-hot encoding of $y$ means that probabilities assigned to all $y' \neq y$ are penalized equally\Omit{ because no $y'$ is the true observed value for the current sequence}.

However, in the presence of constraints, there are really \textit{three} categories of predictions: the partitions previously mentioned. Having a three-level loss function that punishes the partition of illegal predictions more than the legal-but-incorrect prediction could be interpreted in the vein of methods that artificially increase the training-set size, such as left-right reversing of images of scenes in a classification task where it is assumed \textit{a priori} that orientation could not possibly affect the classification. These methods are most effective in situations like images or trees, where the input is high-dimensional but lies on a lower-dimensional manifold. In our work, the logical machine provides some feedback to the NAM that even though certain sequences were not actually in the training data, they have the possibility of existing, while others do not. 

\Omit{\begin{changebar}
The approach taken here does not alter the target $y$ from its one-hot encoding, but instead
adds a separate term to the loss. 
\end{changebar}}
The objective that the NAM optimizes can be written as follows:
\begin{equation} \label{eq:loss}
\textit{Loss} = L_{\mathit{xe}} + \Sigma_{i} \lambda_i L_{c_i}
\end{equation}
where $L_{\mathit{xe}}$ is the traditional cross-entropy loss function and $L_{c_i}$ is the additional penalty for violating constraint $c_i$, whose magnitude is controlled by $\lambda_i$. We say that Equation \ref{eq:loss} defines a \textit{three-level loss function} because predictions that are both wrong and illegal are penalized by both terms, while predictions that are wrong but legal are only penalized by the first. The tradeoff between NAMs learning the specific $(x_i,y_i)$ training sequences versus the constraint more generally\textemdash without caring \textit{which} legal sequences are more realistic\textemdash can be controlled with the hyperparameter $\lambda_i$.

\paragraph{Algorithms.}
The algorithms for training and generation are given as
Algs.~\ref{algtrain} and \ref{alggen}, respectively.
\begin{changebar}[6pt]
During training, the trees in the corpus are traversed and the NAM's
parameters are updated via stochastic gradient descent. Generation
then samples trees from the learned model.
\end{changebar}
\begin{algorithm} \label{algtrain}
trees := SequentializeTrainingTrees()\;
  \Repeat{all trees are processed}{
    tree := nextTrainingTree(trees)\;
    trueNonterminal := curNonterminal\;
    trueOp := curOp
    
      \Repeat{all nonterminals in tree are processed}{
        predictedOp := ChooseOperator(trueNonterminal, curContext)\;
        gradients := Loss(trueOp,predictedOp)\;
        UpdateWeights(gradients)\;
        (trueNonterminal,trueOp) := next position in left-right threading
                                   of sequentialized tree\;
        curContext := update curContext according to the
                    values of the attributes at trueNonterminal\;
         }
   }
\caption{Training}
\end{algorithm}
\begin{algorithm} \label{alggen}
\begin{changebar}[6pt]
tree := Root({[hole]})\;
\end{changebar}
curFocus := tree.child{[1]}\;
curNonterminal := nonterminal of curFocus\;
curContext := the values of the attributes at curFocus\;
\Repeat{there are no more holes}{
     op := ChooseOperator(curNonterminal, curContext)\;
\begin{changebar}[6pt]
     Insert op([$\textrm{hole}_1$], $\ldots$, [$\textrm{hole}_{\mathit{arity}(\mathrm{op})}$]) into tree at curFocus\;
\end{changebar}
     curFocus := next hole in preorder after curFocus\;
     curNonterminal := nonterminal of curFocus\;
     curContext := update curContext according to the
                   values of the attributes at curFocus\;
}
\caption{Generation}
\end{algorithm}

\begin{figure}
	\centering
	\begin{minipage}{.23\textwidth}
	\centering
	\includegraphics[width=\linewidth]{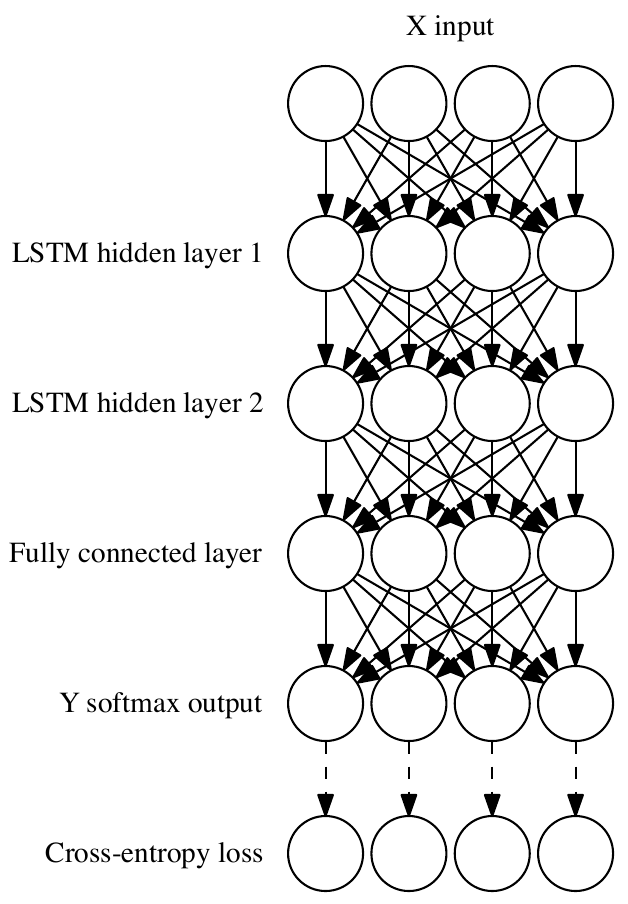}
	\end{minipage}%
	\hfill%
	\begin{minipage}{.3\textwidth}
	\centering
	\includegraphics[width=1\linewidth]{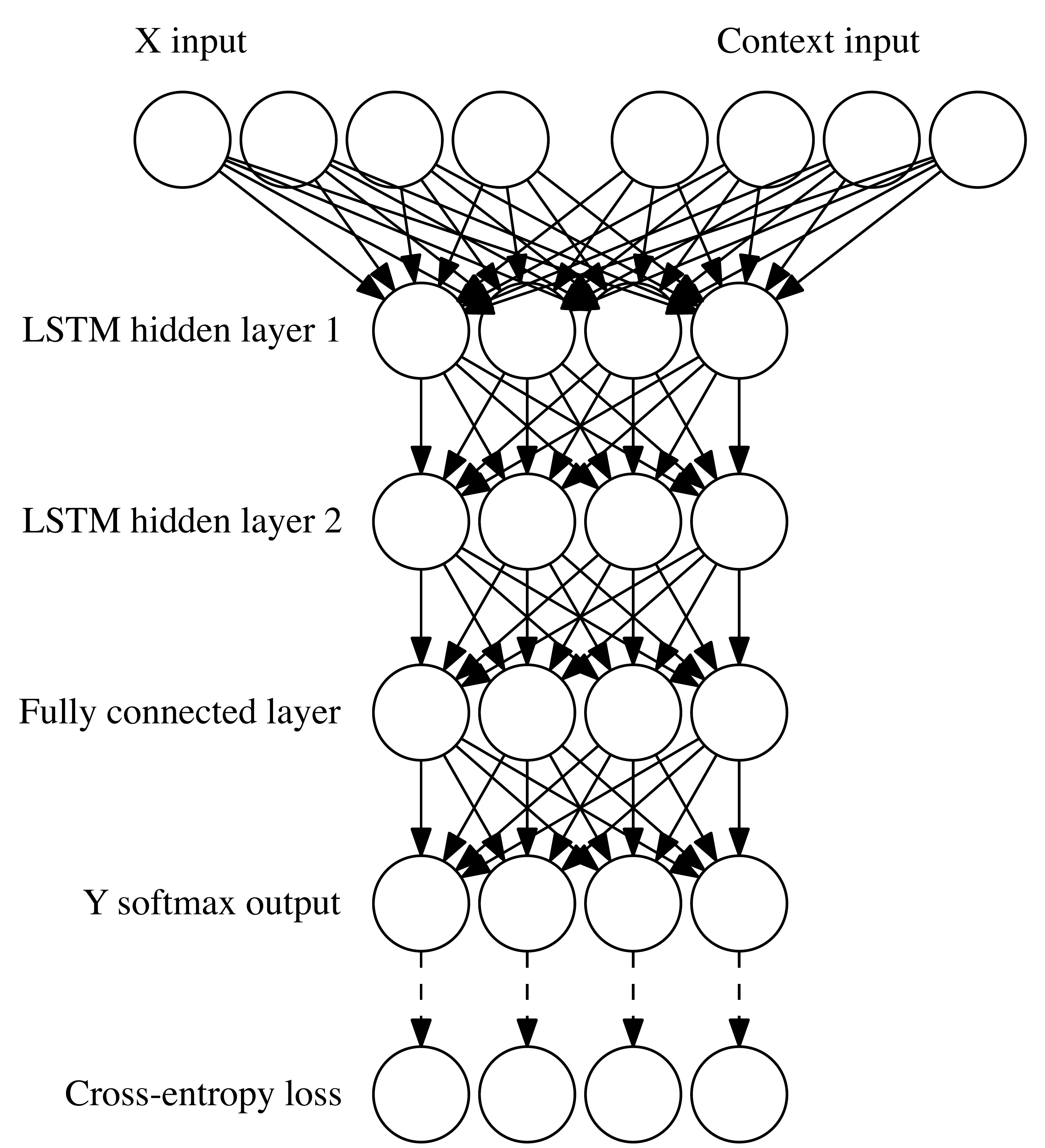}
	\end{minipage}%
	\hfill%
	\begin{minipage}{.45\textwidth}
	\centering
	\includegraphics[width=\linewidth]{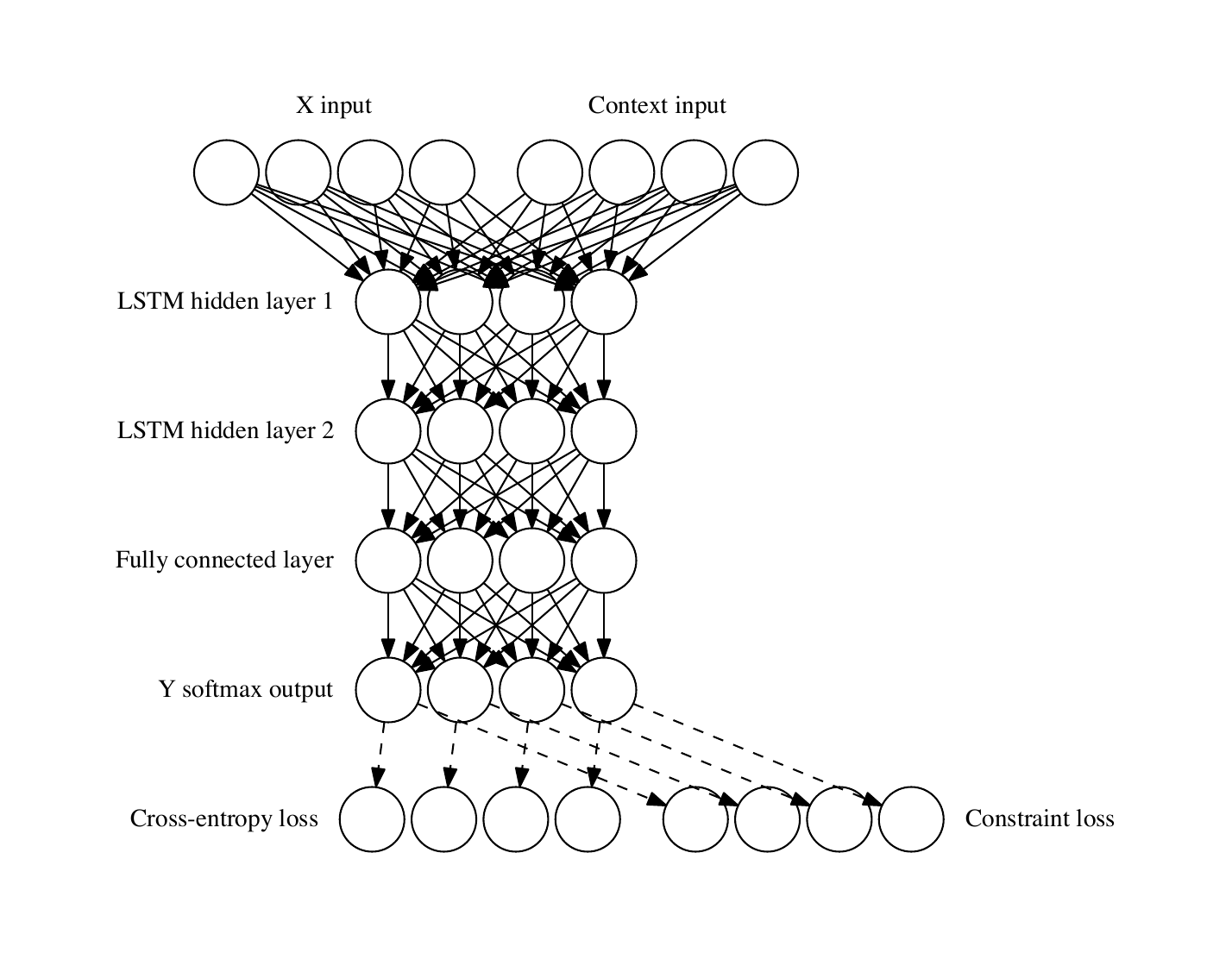}
	\end{minipage}
	
	\caption{(a) Vanilla RNN (b) NAM with context (c) Full NAM}
	\label{fig:rnns}
\end{figure}


\section{Experiments}
\label{ssec:Experiments}

All of our experiments used the following models: a vanilla RNN, a NAM with just the augmented input, a NAM with just the new objective function, and the full NAM with both. The chosen version of RNN is the Long Short Term Memory (LSTM) cell \cite{hochreiter1997long}, which has been favored in recent literature for its ability to learn long-term dependences, although the NAM framework is general to any type of RNN cell \cite{margarit2016batch,stollenga2015parallel,sutskever2014sequence}. For the entire experiment, two stacked LSTMs are used, the number of neurons in each layer is $200$, and backpropagation through time is truncated after $50$ steps. The Adam optimizer \cite{kingma2014adam} is used with learning rate $.001$, dropout \cite{srivastava2014dropout} with a keep probability of $.9$ is applied to all layers except the softmax output, and both $L_1$-norm and $L_2$-norm regularization is applied to weights but not biases in all matrices with $\lambda_{L_1} = \lambda_{L_2} = .0001$. The NAM's $\lambda_i$ values were set to $.1$, chosen so that the order of magnitude of the gradients for each term were roughly equal. As \textit{de novo} generation is the goal in mind, all models were trained until their generation performance no longer improved as measured by the evaluation criteria discussed below.

\begin{changebar}
An artificial corpus created for the work here is a set of 1,500 simple C
programs containing elementary arithmetic operations, variable
manipulations, and control-flow operations, 15\% of which were held out for testing. There are an average of
7.01 unique variables, 3.29 unique types, 6.47 procedures, and 101.82 lines of code per
program, providing a challenge for both constraints by having numerous declarations, multiple types in the same program, and changing scopes. The full corpus is available in the supplementary material.
Programs in the corpus are translated to an abstract syntax
tree (AST) in the C Intermediate Language (CIL) \cite{necula2002cil}.
The set of these ASTs can be interpreted as a sample from an
$L$-attributed attribute grammar $G$.
\end{changebar}

\subsection{Evaluation criteria}
Our experiments were designed to answer the following questions:
\begin{enumerate}[label=\Alph*.]
  \item
    \label{It:QualityOfSimulatedSamples}
    What is the quality of simulated samples?
  \item
    \label{It:RepresentTrainingData}
    How well do the models represent the training data?
  \item
    \label{It:ConstraintViolations}
    At what rate are constraints violated while sampling?
\end{enumerate}
Methods for evaluating the quality of the learned model in generation
tasks can be more subjective than in prediction tasks, where
performance on held-out test sets is relevant.
\begin{changebar}
In our case, however, we can make various measurements of error rates
when internal nodes are generated, as well as test whether a
generated tree satisfies the constraint as a whole, which provides
an overall measure of success.
\end{changebar}

Three measurements are used throughout
\begin{enumerate}
  \item
    \label{It:AbilityToLearnCorpus}
    The ability to learn the corpus as measured by the average negative
    log-likelihood of the training samples under each model.
  \item
    \label{It:IncorrectPreditions}
    The number of predictions made that violate the constraint while
    generating new samples.
  \item
    \label{It:TreesThatSatisfyConstraint}
    \begin{changebar}
    The number of trees that are entirely legal under the specific constraint under consideration. (In
    each generated tree, one constraint violation is sufficient to make
    the whole tree illegal.)
    \end{changebar}
\end{enumerate}

\Omit{
\begin{figure}
\begin{footnotesize}
\begin{minipage}{.48\textwidth}
\begin{lstlisting}[frame=tlrb]{name}
int var0;
...
int main() {
	...
	var3++;
	...
} 
\end{lstlisting}
\end{minipage}%
\hfill%
\begin{minipage}{.48\textwidth}
\begin{lstlisting}[frame=tlrb]{name}
int var0 = 0;
...
float var4;
...
int  main() {
	...
	var4 = var4 + var0;
	...
} 
\end{lstlisting}
\end{minipage}

\end{footnotesize}

\caption{Examples of violations of each constraint made while generating a program. Left: declared-variable violation. Right: type violation.}
\label{fig:exemplarmistakes}

\end{figure}
}
\subsection{Declared-variable constraint} \label{ssec:declared}
Our first experiment imposed the constraint that every variable used must be 
declared (see columns 2-5 of Fig. \ref{table:results}, and Fig. \ref{fig:trainingllhood}). 
As shown by Fig. \ref{table:results}, even though the vanilla RNN gets to 
see the whole tree prior to the node requiring a prediction, it still makes many mistakes. 
For comparison, a stochastic context-free grammar that has been given the same 
augmented input is shown in Fig. \ref{fig:trainingllhood}.
\begin{changebar}
Since it now includes the context vector, it is referred to hereafter as a stochastic grammar with context (SGWC). This model will never choose to 
use variable $v$ if the element of the context vector corresponding to $v$ is not set to 1,
and thus never violates the constraint.
\end{changebar}
However, it does 
not specialize to the corpus very well, motivating the use of neural models that
capture richer patterns in the training-set. The
NAM's modified objective function offers a modest improvement,
but augmenting the input with the context vector provides a much more
significant improvement. Moreover, the latter effect does not dominate
the former: the full NAM with both improvements performs the best
overall.

\Omit{Fewer constraint violations and more trees that are ultimately
accepted are straightforward improvements. }Some insight into how the
models differ can be gleaned from the average negative
log-likelihoods (see Fig.~\ref{fig:trainingllhood}).
\Omit{Lower values mean that the probabilities of the
specific trees in the training-set are higher.}
\begin{changebar}
As expected, the
NAM's loss term acts as a regularization term, and even though
training-set trees are less likely, the result is improved
generation ability (see Fig.~\ref{table:results}).
\end{changebar}
The better generation ability strongly
suggests that the excess fitting that the vanilla RNN did to the
training-set compared to the NAM w/ 3-level loss is best described as
overfitting.

The average number of unique variables and average number of
procedures in each generated program gives one measurement of each
model's fidelity to the training corpus. Training-set trees varied,
but averaged 7.0 unique variables and 6.5 procedures. The
vanilla RNN uses more unique variables and has fewer procedures than
programs in the training-set, corroborating the likelihood numbers'
indication that they did not learn the corpus as well as the other
models. The NAM with context and the NAM with both
improvements yielded samples that resembled the corpus much more
faithfully by these measurements (see Fig.~\ref{table:results}).

Augmenting the input with the context vector makes the representation
of the input at each step \textit{much} richer. Thus, the NAM
with context is able to learn more valuable patterns of the training
data that exist in this higher-dimensional space. The number of legal
trees increases 2.8-fold over the baseline vanilla RNN. This drastic
improvement still has some degree of overfitting that can be
alleviated with the regularizing loss term. The full NAM thus has
a slightly higher average negative log-likelihood than the one without
the regularization, but it has the best results during generation (see
Fig.~\ref{table:results})

\begin{figure}
\centering
\resizebox{\textwidth}{!}{
\begin{tabular}{||c || c | c | c | c || c | c | c | c ||}
  \hhline{|t:=========:t|}
        & \multicolumn{4}{c||}{Declared-variable constraint} & \multicolumn{4}{c||}{Typesafe-variable constraint} \\
  \hhline{||~||====#====||}
  Model & Avg.  & Avg.   & Constraint & Legal & Avg.  & Avg.   & Constraint & Legal \\
        & Vars. & Procs. & Violations & Trees & Vars. & Procs. & Violations & Trees \\
  \hhline{||=#=|=|=|=#=|=|=|=||}
  Vanilla RNN & 8.5 & 4.1 & 9426 & 187  & 8.4 & 4.3 & 7707 & 116 \\
  \hline
  NAM w/ 3-level loss & 8.2 & 4.3 & 8119 & 203  & 8.0 & 4.4 & 6673 & 177 \\
  \hline
  NAM w/ context & 6.7 & 6.7 & 2105 & 532  & 6.6 & 6.7 & 902 & 665 \\
  \hline
  NAM w/ both & 6.8 & 6.7 & 1846 & 582  & 6.5 & 7.0 & 697 & 674 \\
  \hhline{|b:=========:b|}
\end{tabular}
}

\caption{Characteristics of generated trees under the two constraints.
  (1,000 trees were generated separately for each of the two constraints.)
}

\label{table:results}
\vspace{-0.5ex}
\end{figure}

\begin{figure}
\centering
\begin{footnotesize}

\begin{tabular}{|| c || c || c ||}
\hhline{|t:===:t|}
Model & Declared-variable (train/test) & Typesafe-variable (train/test)\\
\hhline{||=#=#=||}
SGWC & 1.856/1.813 & 1.406/7.178\\
\hline
Vanilla RNN & .231/.253 & 1.366/1.425\\
\hline
NAM w/ 3-level loss & .246/.257 & 1.375/1.471\\
\hline
NAM w/context & .181/.188 & .782/.779\\
\hline
NAM w/both & .194/.208 & .819/.794\\
\hhline{|b:===:b|}
\end{tabular}
\caption{Average per token negative log-likelihood of the training data and testing data\Omit{ under the two constraints}.}
\label{fig:trainingllhood}
\end{footnotesize}
\vspace{-2.0ex}
\end{figure}

\subsection{Typesafe-variable constraint}
The same relative performance is seen when we work with the second constraint\Omit{a different
constraint, that of only using variables that agree in type with the
requirements of their position in the tree}. The SGWC generalizes to the test set 
especially poorly in this case, because the context vector varies more and thus there 
are rare or completely novel situations that the SGWC struggles with. The vanilla RNN 
is the same model as in Section \ref{ssec:declared},
because it does not take the constraint into consideration in any
way. The raw number of violations is lower in this setting because the
tests for this constraint occur less frequently: not all variable uses
involve multiple types that must agree. Even so, the simpler models
produce fewer legal trees than in the experiment in Section
\ref{ssec:declared}.\Omit{ The context vector appears to help the NAM
learn more about the underlying probability distribution of the input,
because the training data is more likely under the models utilizing
the context vector than under those that do not (see Fig.~\ref{table:results}).}


\section{Related Work}

The problem of corpus-driven program generation has been studied
before~\cite{ICML:MT14,raychev2014code,nguyenFSE13,nguyenICSE15,bielik2016phog}.
Statistical models used in this task include $n$-gram topic
models~\cite{nguyenFSE13}, probabilistic tree-substitution
grammars~\cite{idiomsFSE14}, a generalization of probabilistic
grammars known as probabilistic higher-order
grammars~\cite{bielik2016phog}, and recurrent neural
networks~\cite{raychev2014code}. The most closely related piece of
work is by Maddison and Tarlow \cite{ICML:MT14}, who use log-bilinear
tree-traversal models, a class of probabilistic pushdown automata, for
program generation. Their model also addresses the problems of
declarations of names and type consistency, and they use ``traversal
variables'' to propagate information from parts of the
already-produced tree to influence what production is selected at a
node.  However, the state-transition function of their generator
admits a simple ``tabular parameterization,'' whereas memory updates
in our approach involve complex interactions between a neural
and a logical machine. Also, their training process does not have an
analog of our three-valued loss function. 

Program generation is closely related to the problem of {\em program
  synthesis}, the problem of producing programs in a high-level
language that implement a user-given specification. A recent body of
work uses neural
techniques~\cite{parisotto2016neuro,murali2017bayesian,balog2016deepcoder}
to solve this problem.
Of these efforts, Balog et al.~\cite{balog2016deepcoder} and Murali et
al.~\cite{murali2017bayesian} use combinatorial search, guided by a
neural network, to generate programs that satisfy language-level
constraints. However, this literature has not studied neural
architectures whose training predisposes them toward satisfying such
constraints.


The work presented here can be related to several key concepts in the
theory of grammars.  Breaking down the generation of a tree into a
series of non-terminals, terminals, and production rules is the same
methodology used with stochastic context-free-grammars.  As an
automaton that sees an input stream that contains occurrences of
``pop'' and produces an output, a NAM is a form of transducer, namely a
visibly-pushdown transducer \cite{ICALP:RS08}.

\Omit{There are also connections between the approach we have used and
methods that take advantage of other special structures. For images,
models that incorporate the grid-like structure such as Spatial LSTMs
or those that explicitly force the image gradient to be smooth take
advantage of relevant structural information---namely, grid
structure---in that domain.}
Neural stack machines like those in \cite{NIPS2015_5648} augment an RNN
with a stack, which the RNN must learn how to operate through differentiable
approximations.
\begin{changebar}[6pt]
In contrast, a NAM only needs to learn how to make use of data values
generated by the logical machine, rather than additionally needing to
learn how to mimic the logical machine's operations.
\end{changebar}

The new term that was introduced in the objective function can be
thought of as a way to perform regularization. Many attempts at
customized regularization have been demonstrated \cite{chien2016bayesian,bai2014sae}. 
Our regularization term allows NAMs to learn the set of all legal
production rules without penalty, but regularizes the learning of the
specific singleton relative to the set of legal production rules.






\section{Discussion}

Learning to generate sequences with strong structural constraints would ideally be as easy as presenting an RNN exclusively with sequences that are members of the constrained space. 
\begin{changebar}
As our experiments show, this can be difficult to achieve in practice.
\end{changebar}
In some cases, though, aspects of the structure can be explicitly represented. NAMs provide a framework for incorporating the knowledge of these constraints into RNNs. They significantly outperform RNNs without the constraint when trained on the same data.

We have demonstrated the utility of NAMs for two $L$-attributed AG
problems. The work here allows for the possibility of creating a
\textit{generator} of NAM systems: from a specification of a desired
constrained language, one could generate the corresponding form for
training.  Moreover, these are just two of the possible types of
mistakes that would prevent a program from passing a C compiler's many
checks.
A topic for future work is to incorporate enough C
constraints so that generated programs would have a high probability
of being compilable.

Because other sequences have a natural underlying parse tree and
associated constraints that can be expressed using an $L$-attributed
AG, another topic for future work is to explore the application
of NAMs to other structured sequences, such as proof trees of a logic.


\bibliographystyle{unsrt}

\end{document}